\documentclass[11pt]{article}

\usepackage[english]{babel}
\usepackage[utf8]{inputenc}
\usepackage[usenames,dvipsnames]{color}

\usepackage[a4paper,left=2.65cm,right=2.65cm,top=2.5cm,bottom=3cm,
bindingoffset=0mm]{geometry}

\usepackage{amsthm,amsmath}
\usepackage{amssymb}

\usepackage{graphicx}

\usepackage{enumitem}

\usepackage{iclr2021_conference}

\title{Improving Simulations with Symmetry\\ Control Neural Networks}

\author{Marc Syvaeri${}^{1,2}$ \& Sven Krippendorf${}^{1}$  \\[0.1cm]
${}^{1}$ Arnold-Sommerfeld-Center \qquad \qquad\qquad${}^{2}$ Max-Planck-Institut f\"ur Physik\\
Ludwig-Maximilians-Universit\"at \qquad \qquad \;\,F\"ohringer Ring 6\\
Theresienstr.~37\qquad \qquad\qquad \qquad\qquad \; \;\;\;\;\,80805 Munich, Germany\\
80333 Munich, Germany \\[0.1cm]
\texttt{\{marc.syvaeri,sven.krippendorf\}@physik.uni-muenchen.de} \\
}

\iclrfinalcopy 
\begin{document}
\vspace*{-2.cm}

\begin{flushright}
  {\small
  LMU-ASC 11/21\\
  MPP-2021-67
  }
\end{flushright}
\vspace*{0.5cm}
\maketitle
\begin{abstract}
The dynamics of physical systems is often constrained to lower dimensional 
sub-spaces due to the presence of conserved quantities. Here we propose a method 
to learn and exploit such symmetry constraints building upon Hamiltonian Neural 
Networks. By enforcing cyclic coordinates with appropriate loss functions, we 
find that we can achieve improved accuracy on simple classical dynamics tasks.  
By fitting analytic formulae to the latent variables in our network we recover 
that our networks are utilizing conserved quantities such as (angular) momentum.
\end{abstract}
\section{Introduction}
\begin{figure}[h!]
 \includegraphics[width=1\textwidth]{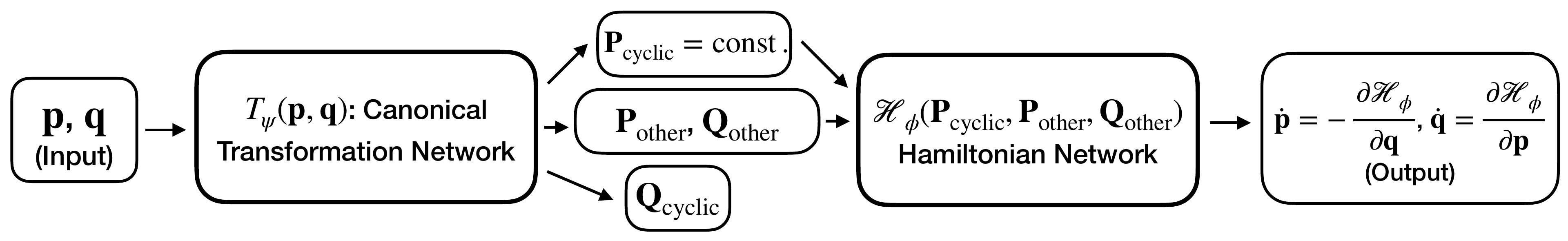}
 \caption{Structure of symmetry control networks: The network $T_{\psi}$ 
transforms the input coordinates of phase space $({\bf p}, {\bf q})$ to 
canonical coordinates where some coordinates are forced to be cyclic. These 
coordinates are used as input to the Hamiltonian network $H_{\phi}$. The output, 
the time derivatives of our initial coordinates, is calculated from the 
Hamiltonian using auto-differentiation.}\label{fig:hlayout}
\end{figure}

For accurate simulations, building in a bias to deep neural networks has been a 
key mechanism to achieve extra-ordinary performance. An example for such a bias 
is to learn a motion which is conserving energy as performed in the context of 
Hamiltonian Neural Networks (HNNs)~\citep{DBLP:journals/corr/abs-1906-01563}. 
More generally speaking, possible dynamics are constrained due to symmetries of 
the system and they take place on a sub-space of available phase space 
(e.g.~surfaces of constant energy and total angular momentum).
Whereas energy has been built into the functional biasing of neural networks, 
further symmetries are at this moment enforced by hand which requires domain 
knowledge of the system. Here we describe a method on how to automatically 
search for and use such symmetries in neural networks used for simulations. In 
addition this method turns out to reveal -- at this stage for simple systems -- 
analytic expressions for the conserved quantities.
Our method is based on enforcing the network to perform canonical coordinate 
transformations in a first network $T_\psi$ which enforces some of the latent 
coordinates to become constant (cyclic coordinates in the language of 
Hamiltonian dynamics). From these new coordinates the network then predicts the 
Hamiltonian underlying the dynamics of this system with a second network 
$H_{\phi}.$ The structure of our networks -- to be described more accurately 
later on -- can be found in Figure~\ref{fig:hlayout}.

We demonstrate in our 
experiments that multiple conserved quantities can indeed be automatically 
identified with this method, we find improvements in the predictions of our 
dynamics in comparison with HNNs, and we can find analytic expressions for the 
conserved quantities.

\section{Theory}
When motion is taking place on a sub-space of phase space, one can reduce the 
number of coordinates describing the system. Every independent conserved 
quantity reduces the number of coordinates needed to describe a system until all 
coordinates are constrained (integrable systems). 
Even though we do not know apriori the conserved coordinates, Hamiltonian 
mechanics provides general conditions for such coordinates. In particular the 
time derivative of any function on phase space, described by positions ${\bf q}$ 
and momenta ${\bf p}$ of our $N$ particles in $d$ spatial dimensions, is given 
by 
\begin{align}
 \frac{d g(\bf{q},\bf{p})}{d t}=\sum_{i=1}^{N\cdot d}\frac{\partial g}{\partial 
q_i}\frac{dq_i}{dt}+\frac{\partial g}{\partial p_i}\frac{dp_i}{dt}=\lbrace 
g,\mathcal{H}\rbrace~.
 \label{eq:timederivative}
\end{align}
In the last step we have used the Poisson brackets and Hamilton's equations 
which are given as:
\begin{align}
	\lbrace f,g\rbrace:= \sum_{i=1}^{N\cdot d} \frac{\partial f}{\partial 
q_i} \frac{\partial g}{\partial p_i} - \frac{\partial f}{\partial p_i} 
\frac{\partial g}{\partial 
q_i}\,,\qquad\frac{d\bf{q}}{dt}=\frac{\partial\mathcal{H}}{d\bf{p}}=\lbrace {\bf 
q},\mathcal{H}\rbrace\,,\qquad	\frac{d{\bf 
p}}{dt}=-\frac{\partial\mathcal{H}}{d\bf{q}}=\lbrace 
\bf{p},\mathcal{H}\rbrace\,.
	\label{eq:heqns}
\end{align}
Here we are interested in enforcing coordinate transformations $T$ which enforce 
that at least one of the new momenta ${\bf P}$ is conserved
\begin{align}
	0=\dot{P}_i=-\frac{\partial \mathcal{H}}{\partial Q_i}=\lbrace 
P_i,\mathcal{H}\rbrace\,.
	\label{eq:conservedconstraint}
\end{align}
We are interested in a particular type of coordinate transformations, namely 
canonical transformations which are which are transformations that leave the 
structure of the Hamiltonian equations~\eqref{eq:heqns} and in particular the 
Poisson bracket unchanged (allowing for the evaluation with respect to the input 
variables):
\begin{align}
\nonumber &T:\left({\bf q},{\bf p}\right)\mapsto\left({\bf Q}({\bf q},{\bf 
p}),{\bf P}({\bf q},{\bf p})\right),\\
 &\lbrace f,g\rbrace_{{\bf p},{\bf q}}=\lbrace f,g\rbrace_{{\bf P},{\bf Q}}~,\;\;\;
\mathcal{H}({\bf p},{\bf q})=\tilde{\mathcal{H}}({\bf P}({\bf p},{\bf q}),{\bf 
Q}({\bf p},{\bf q}))~.
 \label{eq:ctransformation}
\end{align}
We enforce this conservation by adding an appropriate loss term corresponding to 
the condition in~\eqref{eq:conservedconstraint}. When such a constraint is 
enforced, the Hamiltonian does not depend on the associated $Q_i,$ i.e.~it 
depends on fewer degrees of freedom and the motion in phase space is restricted 
to a lower dimensional manifold.
Put differently, the cyclic coordinates provide via constraints of the 
type~\eqref{eq:conservedconstraint} additional restrictions on the allowed 
Hamiltonian function which we learn with our symmetry control neural networks.%
(cf.~Figure~\ref{fig:losscontours} shows the explicit constraint from angular 
momentum conservation in a 2-body example in addition to constraints arising 
from satisfying the Hamiltonian equations of motion).

These conditions lead to two additional loss components in addition to the 
standard HNN-loss:
\begin{enumerate}[leftmargin=0.4cm]
 \item The first loss ensures that our Hamiltonian satisfies Hamiltonian 
equations~\eqref{eq:heqns}, which we can ensure as follows:
\begin{equation}
	\mathcal{L}_{\rm HNN}=\sum_{i=1}^{N\cdot d}\left\lVert\frac{\partial 
\mathcal{H}_{\phi}({\bf P},{\bf Q})}{\partial p_i}-\frac{d q_i}{d 
t}\right\rVert_2+
	\left\lVert\frac{\partial \mathcal{H}_{\phi}({\bf P},{\bf Q})}{\partial 
q_i}+\frac{d p_i}{d t}\right\rVert_2\,.
	\label{eq:hnnloss}
\end{equation} 
The time derivatives are provided by the data and the derivatives of the 
Hamiltonian with respect to the input variables can be obtained using 
auto-differentiation. This is the same loss as introduced 
in~\cite{DBLP:journals/corr/abs-1906-01563}.
 \item To ensure that our transformation $T_\psi$ are of the type we are 
interested in (cf.~Eq.~\eqref{eq:ctransformation}), i.e.~our new coordinates 
fulfil the Poisson algebra, we enforce the following loss:
 \begin{equation}
  \mathcal{L}_{\rm Poisson}=\sum_{i,j=1}^{N\cdot d}\left\lVert\lbrace 
Q_i,P_j\rbrace-\delta_{ij}\right\rVert_2+\sum_{i,j>i}^{N\cdot 
d}\left\lVert\lbrace P_i,P_j\rbrace\right\rVert_2+\left\lVert\lbrace 
Q_i,Q_j\rbrace\right\rVert_2~,
  \label{eq:poissonloss}
 \end{equation}
 where in some practical applications we only enforce this loss on $n$ cyclic 
coordinate pairs. The first part of this loss ensures that a vanishing solution 
is not allowed.
 \item Hamilton's equations have still to be satisfied with respect to the new 
coordinates. For the cyclic coordinates we have enforced by our architecture 
that ${\cal H}_{\rm \phi}$ is independent of $Q_i.$ To ensure that $P_i$ is 
actually conserved, we require the following additional loss:
 \begin{align}
   \nonumber\mathcal{L}_{\rm HQP}^{(n)}=&\sum_{i=1}^n \left\lVert {\displaystyle 
\frac{dP_i}{dt}}\right\rVert_2\!+\!\left\lVert\frac{dQ_i}{dt}-\frac{\partial 
\mathcal{H}_{\phi}({\bf P},{\bf Q})}{\partial P_i} \right\rVert_2\\& { 
+\beta\sum_{i=n+1}^{N\cdot d}\left\lVert\frac{dP_i}{dt}+\frac{\partial 
\mathcal{H}_{\phi}({\bf P},{\bf Q})}{\partial Q_i} 
\right\rVert_2+\left\lVert\frac{dQ_i}{dt}-\frac{\partial \mathcal{H}_{\phi}({\bf 
P},{\bf Q})}{\partial P_i} \right\rVert_2~,}
 \end{align}
 where $n$ denotes the number of cyclic variables we are imposing and $\beta$ 
denotes a hyperparameter. For our networks we only constrain the cyclic 
coordinates and use $\beta=0$. The time derivatives can be calculated using 
either expressions in~\eqref{eq:timederivative}.
\end{enumerate}
Our total loss is a weighted sum of these three components:
\begin{equation}
 {\cal L}=\mathcal{L}_{\rm HNN}+\alpha_1\mathcal{L}_{\rm 
Poisson}+\alpha_2\mathcal{L}_{\rm HQP}^{(n)}~,
\end{equation}
where the weights $\alpha_i$ are tuned.

For integrating the solutions in time from our respective symmetry control 
network we use a fourth order Runge-Kutta integrator as 
in~\cite{DBLP:journals/corr/abs-1906-01563} which unlike symplectic integrators 
allows for a comparison with neural network approaches directly predicting the 
dynamics of a system.\footnote{We find that the main numerical inaccuracy in the 
prediction arises from inaccuracies of the Hamiltonian ${\mathcal H}_\phi$ 
rather than the choice of this integrator when comparing it with standard 
symplectic integrators~\citep{rebound}.}

Regarding the efficiency of this approach, there is no additional cost during the inference procedure.
During training our loss requires the additional 
evaluation of already calculated gradients. The number of these terms scales 
cubically with the product of number of dimensions and particles $(N\cdot d)^3$ 
(cf.~Poisson loss). 
This cost is lowered when the conserved quantities are known (e.g.~from prior 
knowledge of the system) as the Poisson loss is automatically satisfied and we 
only invoke an additional loss contribution on the Hamiltonian for each 
conserved quantity.
We note that for cases where the Hamiltonian involves only local interactions an 
evaluation of these constraints on these local interactions is relevant which 
results in a reduction of the relevant additional loss terms to the respective 
local terms, significantly reducing the calculational cost. Hence the cost is not significantly enhanced compared to the HNN case.

\section{Experiments}
Our experiments\footnote{Details on hyperparameters and training can be found in 
Appendix~\ref{app:hyperparameters}.} are designed with the following goals in mind:
\begin{enumerate}[leftmargin=0.7cm]
\item {\bf SCNN:} We want to compare the performance of symmetry control neural 
networks with HNNs and baseline neural networks which directly predict 
$(\dot{\bf q},\dot{\bf p})$. We want to take advantage of unknown underlying 
symmetries while we do not have any domain knowledge of our system. The main 
hyperparameter is the number of conserved quantities.

\item {\bf SCNN-constraint:} We explore whether imposing domain knowledge about 
symmetries improves the performance. This is motivated by the fact that we often 
know about the existence of certain conserved quantities such as (angular) 
momentum.
\end{enumerate}
\begin{figure}
\centering
        
\includegraphics[width=0.185\textwidth]{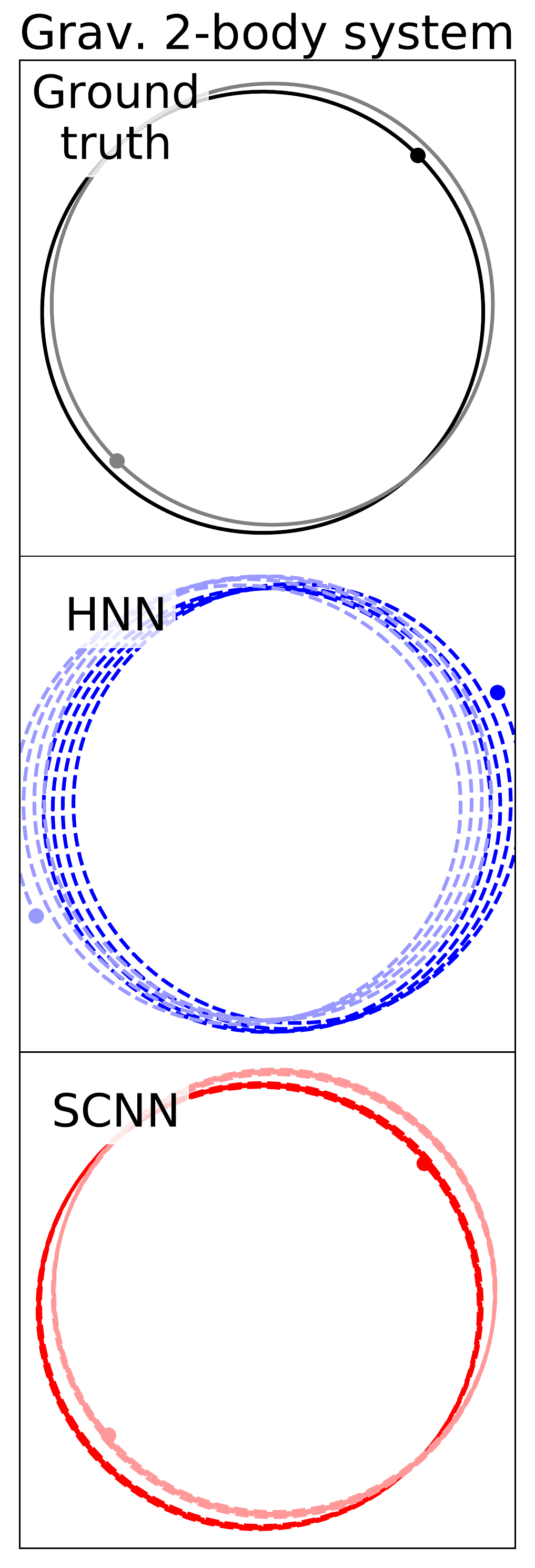}\hspace{0.3cm}
        \includegraphics[width=0.37\textwidth]{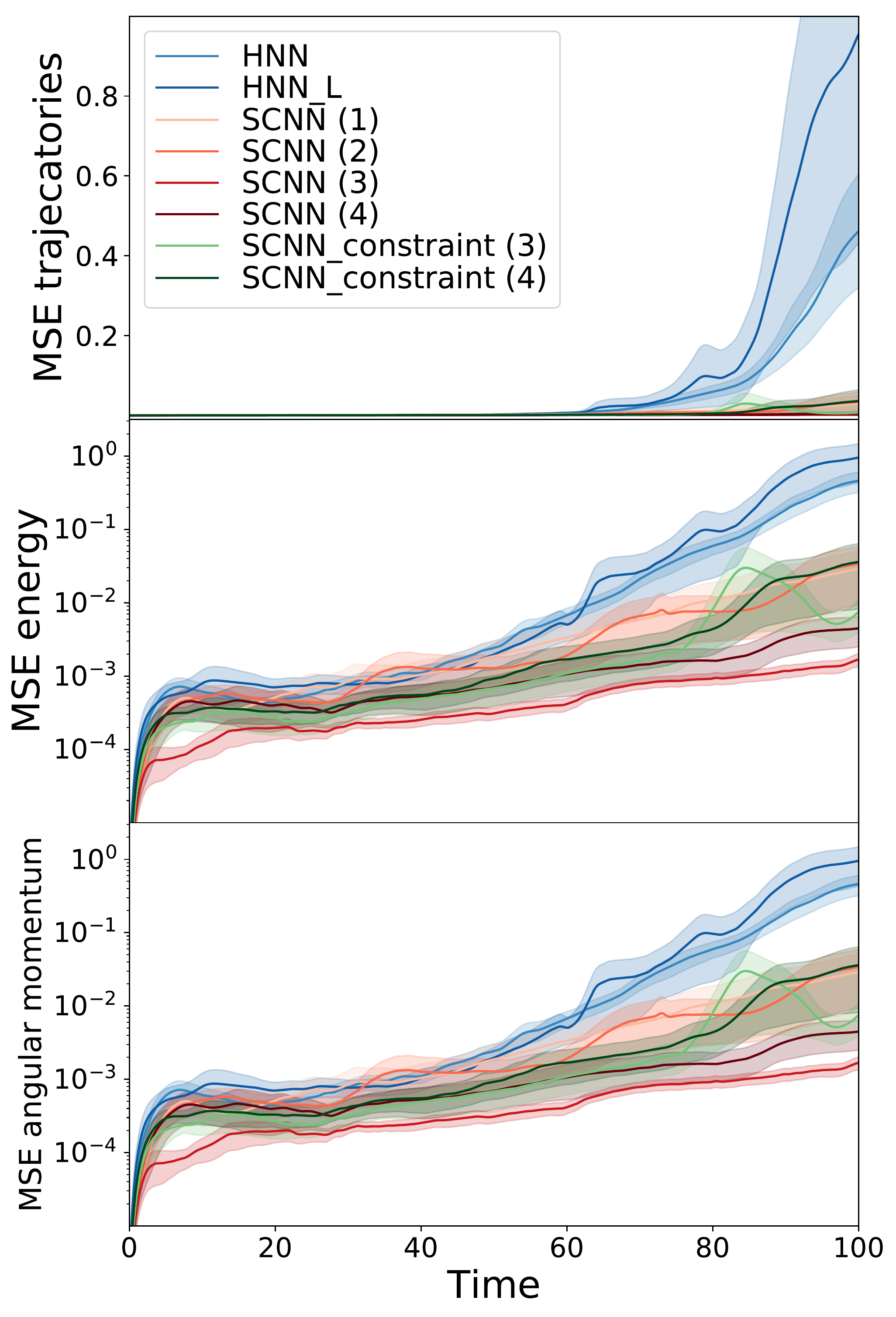}\hspace{0.3cm}
        \includegraphics[width=0.37\textwidth]{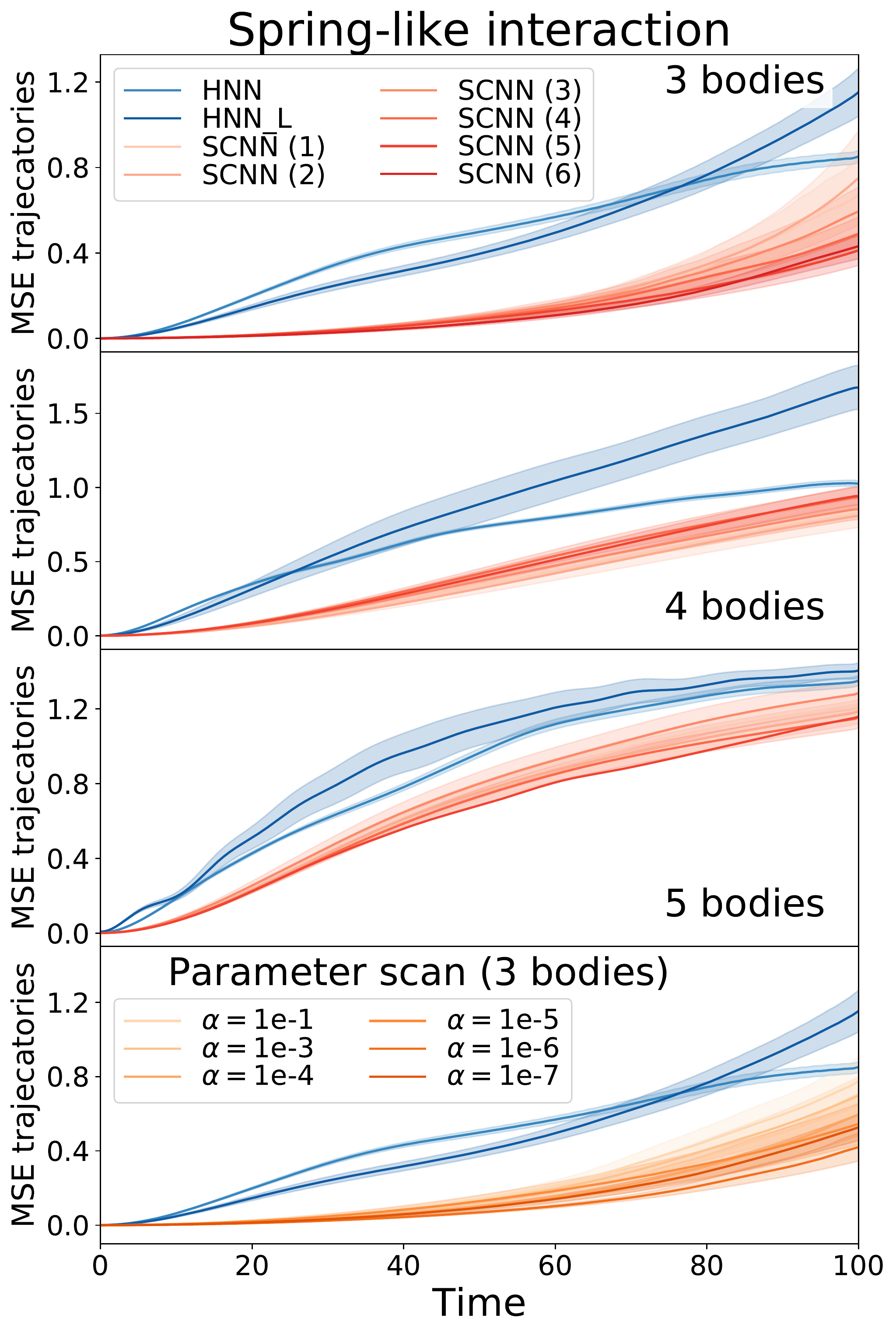}
	\label{fig:results}\vspace{-0.3cm}
	\caption{
	Comparison between two HNN networks as described in the main text and 
SCNN networks with different numbers of conserved quantities. {\bf Left:} Sample 
trajectories of the gravitational two-body problem.
	{\bf Middle:} Comparison of the mean-squared deviation of the 
trajectories, the conservation of energy and angular momentum for the 2-body 
problem. {\bf Right:} Comparison of the mean-squared deviations for various 
spring-like interactions. The bottom plot shows the well-behaved dependence on 
the loss prefactor $\alpha=\alpha_i.$	For all evaluations of the trajectories 
we show the mean and single standard deviation over $10$ different 
initializations evaluated on $100$  trajectories. $\alpha_1=\alpha_2=10^{-n}$ 
where $n$ is the number of constraint conserved quantities.}
\end{figure}

We test these architectures on several different physical systems: the 
gravitational two-body problem, an $n$-body system with spring-like 
interactions, an electrically charged particle in a magnetic field, a spherical 
pendulum, and a double pendulum. We describe the hyperparameter dependence of 
our architecture on the former two in more detail and report only the results of 
our hyperparameter scan for the latter three.
The spring-like interaction contained between 3 and 5 particles each interacting 
with the force ${\bf F^{ij}}= - k~\left({\bf x}^i-{\bf x}^j\right)$. 
The gravitational force is described by ${\bf F^{ij}}= - k~\frac{({\bf x}^i-{\bf 
x}^j)}{|{\bf x}^i-{\bf x}^j|^3}$. 

To understand the effect of the number of conserved quantities and the loss 
prefactors $\alpha_i$, we performed a parameter scan for the number of conserved 
quantities and these prefactors. To compare the precision of the networks we 
computed the particle trajectories of $100$ different system initializations and 
compared them to the ground truth over $100$ timesteps. In addition, we check 
the conservation of energy and angular momentum for the gravitational 2-body 
problem. Our networks are trained on datapoints corresponding to the first $50$ 
timesteps which allows us to compare the generalization of our networks on 
unseen regions of phase-space. The results are summarized in 
Figure~\ref{fig:results}.
 Overall, these results show that the SCNN learns a much more stable and accurate 
Hamiltonian and we find that the SCNN-constraint networks are not able to 
outperform the SCNN networks. The baseline network is omitted from these plots 
as its performance is significantly worse. The HNN network has the same 
complexity as our ${\cal H}_\phi$ network with two dense layers and the HNN\_L 
network corresponds to a complexity of five dense layers, i.e.~the combined 
complexity of our $T_\psi$ and ${\cal H}_\phi$ network. 
The loss prefactor dependence is such that the more conserved quantities we use, 
the smaller the prefactors have to be which is due to the increase in loss terms 
contributing.

In the setup with a charged particle in a magnetic field, a spherical pendulum, 
and the double pendulum
the improved performance of our SCNN(-constraint) networks compared to HNN is 
confirmed. Table~\ref{tab:MSEtable} shows the mean-squared error predictions 
which we obtain after $100$ timesteps. A description of these systems can be 
found in Appendix~\ref{app:otherphysicalsystems}.

\begin{table}
\begin{center}
\renewcommand{\arraystretch}{1.25}
\begin{tabular}{l|ccc } 
Experiment &   SCNN & SCNN-constraint  & HNN \\ \hline
    Magnetic field  & $0.164 \pm 0.025$ &  ${\bf 0.033 \pm 0.010}$&$ 0.083 \pm 0.019$\\
    Spherical pendulum  & $0.092 \pm  0.33$ &${\bf 0.055 \pm 0.005}$  &$0.288 \pm 
0.007$\\
    Double pendulum & ${\bf 0.014 \pm 0.004} $& -- & $0.117 \pm 0.012$\\
    \hline
  \end{tabular}
  \end{center}
  \caption{The MSE after 100 time steps for different experiments. For the 
double pendulum we compare the MSE after 20 time steps. The data points are 
collected for 10 different initalizations and the error bars correspond to 
single standard deviation. All SCNN networks in these experiments have two 
conserved quantities. The best performing approach is highlighted.}\label{tab:MSEtable}
  \end{table}

\subsection*{Search for conserved quantities}
The conserved quantities of our SCNN networks can be analysed by fitting a 
low-dimensional polynomial ansatz to the respective network predictions. 
This reveals that our SCNN finds the angular momentum and the total momentum as 
conserved quantities in the gravitational two-body system:
\begin{equation}
\begin{split} 
 P_{c_1}= &
-4.2~p_{x_1}-4.2~p_{x_2}-1.3~p_{y_1}-1.3~p_{y_2}~,\\
P_{c_2}=&-0.9~p_{x_1}-0.9~p_{x_2}
-3.2~p_{y_1}-3.2~p_{y_2}~,\\
 L=&~1.0~q_{x_1}p_{y_1}+0.9~q_{x_1}p_{y_2}+0.9~q_{x_2}p_{y_1}-1.0~q_{x_2}p_{y_2}\\ &
+1.0~q_{y_1}p_{x_1}-0.9~q_{y_1}p_{x_2}-0.9~q_{y_2}p_{x_1}+1.0~q_{y_2
}p_{x_2}~.
\end{split}
\end{equation}
For more sophisticated conserved quantities (i.e.~non-polynomial conserved 
quantities) different ans\"atze seem necessary (some of which are pursued in 
related work 
\citep{pmlr-v80-sahoo18a,cranmer2019learning,wetzel2020discovering}.

\section{Outlook and related work}
Our SCNNs naturally connect with work on inferring dynamics with neural networks 
such as~\citep{battaglia2016interaction} in the same way as HNNs. Natural 
extensions of our current work can include the application on graph neural 
network based approaches~\citep{sanchez2019hamiltonian} and in 
flows~\citep{Toth2020Hamiltonian}. With respect to applications, we particularly 
look forward to applying our new approach for automatically inferring and using 
the symmetries in astrophysical and molecular dynamics settings.
\bibliography{Bib_PNN} 
\bibliographystyle{iclr2021_conference}

\appendix
\section{Details on further physical systems}
\label{app:otherphysicalsystems}
Here we describe the additional three physical systems which we used in our experiments. 
\subsection{Spherical Pendulum}
The spherical pendulum is governed by the following Hamiltonian:
\begin{align}
 \mathcal{H}=\frac{p_{x}^2+p_{y}^2}{2m} - 
m~g~\sqrt{l^2-q_x^2-q_y^2}~,\label{eq:hamiltonian_pendulum_2}
\end{align}
where we take a universal mass $m=\frac{1}{2}$ coupling $g=\frac{1}{2}$ and 
pendulum length $l=1$. 
In this system, the angular momentum in z-direction is conserved:
\begin{align}
L=q_{x} p_y - q_y p_x~.
\label{eq:conservedpendulum}
\end{align}

We sample the starting conditions 
$\left(\bf{q}\left(t=0\right),\bf{p}\left(t=0\right)\right)$ from a uniform 
distribution from the interval $(-1,1)$. We ensure, that the energy is 
well-defined and negative.

For our dataset we utilise 100 initial conditions and generate 500 points on the 
trajectory. These points are take from a time span of 20 which corresponds to a 
time where multiple periodic motions are included. We use a 80:20 split for the 
training and test set.

\subsection{Charged particle in a magnetic field}
The charged particle in a magnetic field, where we choose a circular B-field 
plus an electric field, is governed by the following Hamiltonian:
\begin{align}
 \mathcal{H}=\frac{1}{2m}\left( \left(p_x+q~q_y~ B\right)^2+ \left(p_y-q~q_x~ 
B\right)^2\right) + k~\left(q_x^2 + q_y^2\right)~,
\end{align}
where we take a universal mass $m=\frac{1}{2}$, electric field $k=1$, charge 
$q=1$ and B-field $B=1$. Again angular momentum is conserved.
We sample the starting conditions 
$\left(\bf{q}\left(t=0\right),\bf{p}\left(t=0\right)\right)$ from a uniform 
distribution from the interval $(-1,1)$, We then normalize the vector to unit 
length and then re-scale it uniformly to a vector of length in the interval 
$(0.1,1.0).$
For our dataset we utilise 100 initial conditions and generate 500 points on the 
trajectory. These points are take from a time span of 20 which corresponds to a 
time where multiple periodic motions are included. We use a 80:20 split for the 
training and test set.

\subsection{Double pendulum}
The double pendulum can be described by the following Hamiltonian:
\begin{align}
 \nonumber 
\mathcal{H}=&~\frac{l_2^2m_2p_2^2+l_1^2(m_1+m_2)p_2^2-2m_2l_1l_2p_1p_2 
\cos{\left(\theta_1-\theta_2\right)}}{2l_1^2l_2^2m_2\left(m_1+m_2\sin^2{
(\theta_1-\theta_2)}\right)}\\ 
&-(m_1+m_2)gl_1\cos{\theta_1}-m_2gl_2\cos{\theta_2}~,
\end{align}
where we set all scales to unity, i.e.~$m_1=m_2=l_1=l_2=g=1.$ We sample the starting conditions 
$\left(\bf{q}\left(t=0\right),\bf{p}\left(t=0\right)=0\right)$ from a uniform 
distribution from the interval $(0,2\pi)$. Note, that $\theta_1$ and $\theta_2$ 
are angles and therefore, we have to ensure that they always lie in the 
interval~$(0,2\pi)$. 
For our dataset we utilise 100 initial conditions and generate 500 points on the 
trajectory. These points are take from a time span of 20 which corresponds to a 
time where multiple periodic motions are included. We use a 80:20 split for the 
training and test set.

\section{Hyperparameters}
\label{app:hyperparameters}

We have performed some hyperparameter searches for our symmetry control networks 
on which we give an overview here. 

By varying the hidden layer size between 50-300, we find that there is a minimum 
size of layer size $200$ to find convergence of our networks. We have varied the 
number of hidden layers up to $5$ hidden layers. Two hidden layers are already 
sufficient (tested up to 5 hidden layers), hence we restricted ourselves on 
them. Coarsely speaking, we find that the precise architecture is less relevant 
in our current experiments.

More relevant are the pre-factors in the loss. Depending on the choice, we can 
either force our networks for better performance on the predictions or obtaining 
the conserved quantities. We have used $\alpha_i$ in the range $[10^{-7},1].$ 
For this trade-off, we have optimized the experiments in this paper on the 
particle trajectories.

We find that pytorch's standard orthogonal initialization provides the best 
results out of the standard initializations. We have not observed a large random 
seed dependence.

As datasets we use the nearly circular orbits constructed by 
\cite{DBLP:journals/corr/abs-1906-01563}, but give the whole system a boost in a 
random direction sampled from $\mathcal{N}\left(0,0.1\right)^2$. For the 
spring-like interaction, we initialize the first $n-1$ particles with 
$\left(\mathbf{q},\mathbf{p}\right)$ from 
$\mathcal{U}\left(\left[-1.5,1.5\right]\right)^4$. For the $n$-th particle, we 
use $\left(\mathbf{q},\mathbf{p}\right) = \left(\sum_i \mathbf{q}_i, \sum_i 
\mathbf{p}_i\right)$, which ensures that the center of mass is at rest.

\end{document}